%% file: main.tex
\documentclass[11pt,a4paper,logo,copyright,nonumbering]{deepseek}
\usepackage{fontawesome}
\usepackage{wrapfig}
\input{preamble}
\title{\centering Changing Environments, Grounded Decisions: \textbf{SADrive}, A Real-World Driving Dataset for Scene-Action Association Learning}

\author[]{
Peiwei Chen$^{1}$, Yuanpeng Chen$^{2  \clubsuit}$, Hui Song$^{2}$, Shanhui Mo$^{3}$, Shuang Zhang$^{2}$, 

\vspace{-0.1in}
Tiankun Zhao$^{2}$, Xiangjing An$^{2}$, Zhigang Ling$^{1 \text{\scalebox{0.7}{\faEnvelopeO}} }$, Yaonan Wang$^{1}$

\vspace{-0.1in}
\small
\text{\scalebox{0.7}{\faEnvelopeO}} Corresponding Author \quad $\clubsuit$ Project Leader \\
\vspace{-0.1in}
\small
$^1$School of Artificial Intelligence and Robotics, Hunan University, Changsha, China 

\vspace{-0.1in}
\small
$^2$Go Further. AI  \quad $^3$Independent-Researcher\\

\small
Email: zgling\_hunan@126.com
}

\input{sec/0_abstract}
\begin{document}
\maketitle
\begin{figure}[!htbp]
\centering
\includegraphics[width=0.78\linewidth]{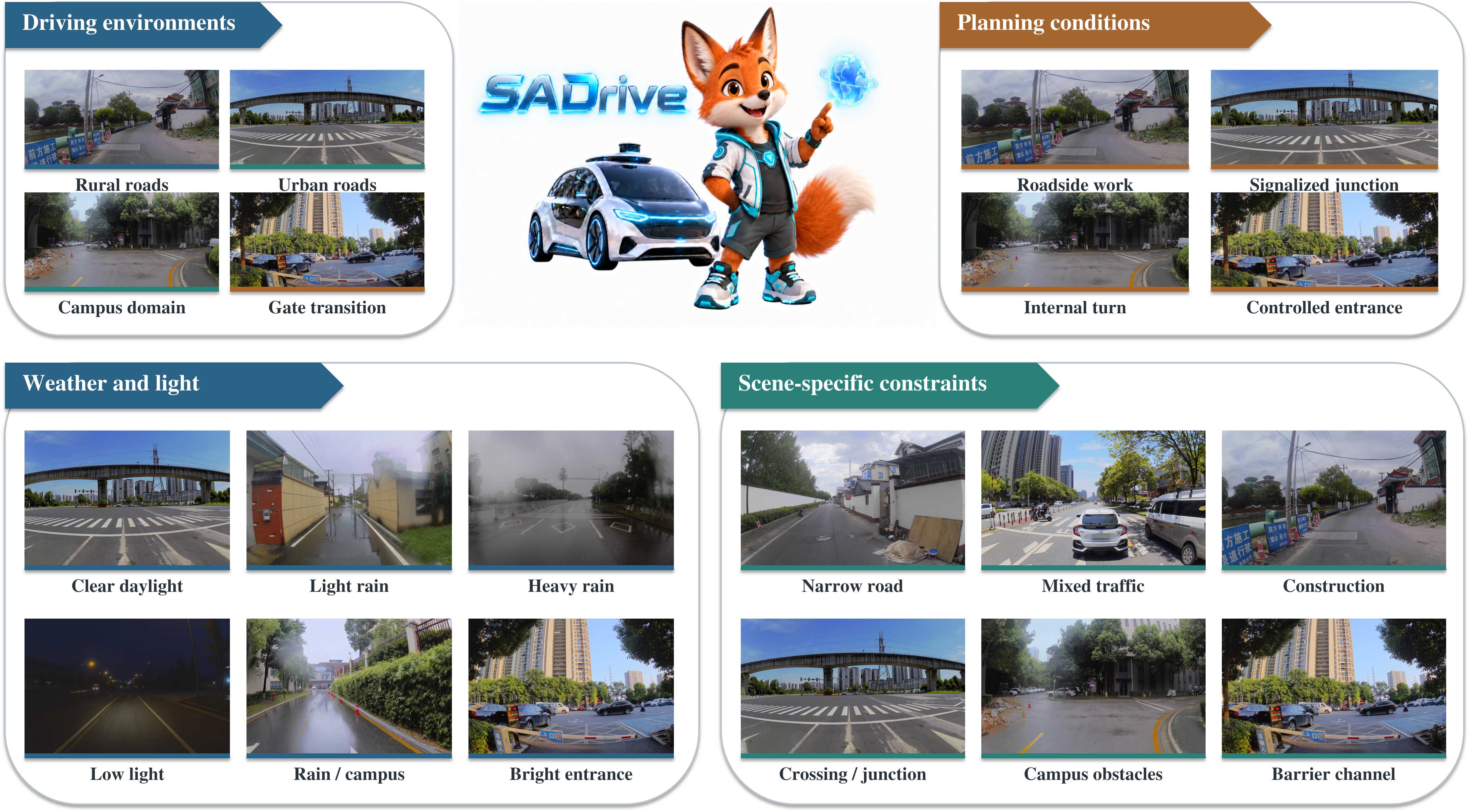}
\caption{\textbf{Overview of the SADrive dataset.}
        The data cover rural roads, urban roads, closed campuses,
        and gate transitions, under different weather and lighting
        conditions. Representative planning situations include
        roadside construction, signalized intersections, and gate
        entrances, as well as narrow roads, mixed traffic, local
        obstacles, and gate passages.}
\label{fig:overview}
\end{figure}
\input{sec/1_intro}

\input{sec/2_related_work}

\input{sec/3_dataset}

\input{sec/4_benchmark}

\input{sec/5_experiments}

\input{sec/6_conclusion}

\input{sec/LLM_Usage_Statement}
\input{sec/X_suppl}
\FloatBarrier
\clearpage
\bibliography{main}
\end{document}

%% file: preamble.tex
\usepackage[authoryear,sort&compress,round]{natbib}
\usepackage{array,multirow}
\usepackage{placeins}
\usepackage{adjustbox}
\ifXeTeX
  \microtypesetup{tracking=false}
  \usepackage{newunicodechar}
  \newunicodechar{–}{\textendash}
  \newunicodechar{’}{\textquoteright}
\fi
\definecolor{groupbg}{RGB}{232,236,250}
\hypersetup{
  colorlinks=true,linkcolor=blue,citecolor=blue,urlcolor=blue,
  pdftitle={Changing Environments, Grounded Decisions: SADrive},
  pdfkeywords={End-to-end autonomous driving, driving dataset, scene-action association, conditional response evaluation, CoCT}
}
\newcommand{\pending}{\textemdash}
\newcommand{\missingimage}[1]{%
  \fbox{\parbox[c][32mm][c]{0.92\linewidth}{%
    \centering Image file missing\\\small\texttt{\detokenize{#1}}}}}
\newenvironment{fittable}{\begin{adjustbox}{max width=\linewidth}}{\end{adjustbox}}
\renewcommand{\today}{}
\author[]{*}
\AtBeginDocument{%
}

%% file: sec/0_abstract.tex
\begin{abstract}
Driving models must adjust their actions as driving conditions change or different environments. Common trajectory error only provides limited insight into these scene--action relationships. We introduce \textbf{SADrive}, a unified data and evaluation foundation for studying scene--action association learning. Unlike existing datasets, SADrive is built from 237 hours of real-world driving and contains 2,942 clips covering rural roads, urban roads, closed campuses, and gate transitions. We further propose two metrics: Conditional Response Difference Error (CRDE) and Cross-Environment Conditional Accuracy (CECA). CRDE measures the error between the predicted response differences and the recorded response differences under changed driving conditions within the same environment. CECA measures agreement with recorded actions under matched driving conditions across environments. The dataset also provides 282,952 SAT(Scene--Action association Token) annotations that explicitly express the links between scene factors and driving actions, enabling the study of scene--action association for learning to plan. 
\end{abstract}

%% file: sec/1_intro.tex
\section{Introduction}

\begin{table}[!htbp]
\centering
\footnotesize
\setlength{\tabcolsep}{3pt}
\renewcommand{\arraystretch}{1.12}
\caption{\textbf{Comparison of dataset coverage and evaluation focus.}
R, U, and P denote rural roads, urban roads, and campus environments.
Plan denotes a ego planning task.
$\checkmark$ indicates explicitly documented coverage, whereas NR denotes coverage or Gate subset size that is not separately reported.}
\label{tab:dataset_comparison}
\resizebox{\textwidth}{!}{%
\begin{tabularx}{\linewidth}{@{}>{\raggedright\arraybackslash}p{0.145\linewidth}>{\raggedright\arraybackslash}p{0.205\linewidth}ccccrc>{\raggedright\arraybackslash}X@{}}
\toprule
\multirow{2}{*}{Dataset} &
\multirow{2}{*}{Reported scale} &
\multirow{2}{*}{Source} &
\multicolumn{3}{c}{Environments} &
\multirow{2}{*}{Gate clips} &
\multirow{2}{*}{Plan} &
\multirow{2}{*}{Evaluation focus}\\
\cmidrule(lr){4-6}
& & & R & U & P & & &\\
\midrule
KITTI\newline\citep{geiger2013vision}
& 14,999 images & Real
& $\checkmark$ & $\checkmark$ & NR
& NR & -- & Object classes \\

nuScenes\newline\citep{caesar2020nuscenes}
& 1,000 scenes & Real
& NR & $\checkmark$ & NR
& NR & -- & Object classes\\

Argoverse 2\newline\citep{wilson2023argoverse}
& 1,000 logs & Real
& NR & $\checkmark$ & NR
& NR & -- & Object classes\\

K-Radar\newline\citep{paek2022k}
& 35k frames & Real
& NR & $\checkmark$ & NR
& NR & -- & Weather / road conditions\\

Ithaca365\newline\citep{diazruiz2022ithaca365}
& 15km route & Real
& $\checkmark$ & $\checkmark$ & $\checkmark$
& NR & -- & Weather / perception \\

\midrule
nuPlan\newline\citep{karnchanachari2024nuplan}
& 1,282h & Real
& NR & $\checkmark$ & NR
& NR & $\checkmark$ & Scenarios / cities\\

NAVSIM v1\newline\citep{dauner2024navsim}
& 103k train + 12k test samples & Real
& NR & $\checkmark$ & NR
& NR & $\checkmark$ & Aggregate PDMS / subscores\\

Bench2Drive\newline\citep{jia2024bench2drive}
& 2M frames; 220 test routes & Sim.
& $\checkmark$ & $\checkmark$ & $\checkmark$
& NR & $\checkmark$ & Five driving abilities\\

WOD-E2E\newline\citep{xu2025wode2e}
& 4,021 segments; 12h & Real
& NR & $\checkmark$ & NR
& NR & $\checkmark$ & ADE / human-preference RFS\\

nuReasoning\newline\citep{huang2026nureasoning}
& 20k clips; 105h; 247k spatial + 57k decision & Real
& NR & NR & NR
& NR & $\checkmark$ & Spatial / counterfactual reasoning\\

\midrule
\textbf{SADrive}
& \textbf{2,942 clips; 473.8k frames;13h} & \textbf{Real}
& $\checkmark$ & $\checkmark$ & $\checkmark$
& 955 & $\checkmark$
& \textbf{Scene--action relationships}\\
\bottomrule
\end{tabularx}
}
\end{table}

Traditional autonomous driving systems make driving decisions through perception, prediction, and planning modules~\cite{liu2025mmtl,li2024bevformer}. In recent years, vision-based end-to-end (E2E) methods have directly linked sensor observations to driving actions. They learn environment representations and planning policies within a unified framework~\citep{jiang2023vad,sun2025sparsedrive}. Multimodal large language models further introduce semantic knowledge and reasoning into driving decisions~\citep{xu2024drivegpt4,xu2024vlm,wang2025alpamayo}. Language explanations, structured question answering, and reasoning supervision also help express the links between environmental factors and driving actions~\citep{tian2025nuscenes,hou2026mmdrive}. These advances support scene--action association learning and raise further evaluation questions. Does a model use the key scene factors it observes as the basis for driving decisions? When their states or relations to the ego path change, do its actions change accordingly?

Evaluating scene--action association learning involves two related questions. First, an E2E model should adjust its actions when key driving conditions change. Second, it should respond correctly when similar passage conditions occur in different environments. The former examines action responses to condition changes. The latter examines the applicability of these associations across environments. Matched comparisons under condition and environment changes are therefore needed to evaluate the correspondence between action choices and passage conditions.

Existing driving datasets and benchmarks support driving research through data collection across environments, scenario classification, and fine-grained evaluation~\citep{caesar2020nuscenes,diazruiz2022ithaca365,karnchanachari2024nuplan,dauner2024navsim,jia2024bench2drive}. However, evaluating scene--action association learning requires a consistent planning interface. It also requires explicit links between key driving conditions and recorded actions, together with paired comparisons under condition and environment changes. Existing benchmarks provide insufficient joint support for these requirements, as shown in Table~\ref{tab:dataset_comparison}. This makes it difficult to establish comparable evaluations of scene--action associations.

To this end, we introduce \textbf{SADrive}, a real-world driving dataset for evaluating scene--action association learning. As shown in Figure~\ref{fig:overview}, SADrive contains 2,942 clips selected from 237 hours of driving records. The clips cover rural roads, urban roads, closed campuses, and gate transitions, totaling approximately 13 hours. Each clip provides observations from 11 cameras(7 surround-view cameras and 4 fisheye cameras), navigation routes, high-level navigation commands, 3D object annotations, semantic occupancy annotations, and ego-vehicle trajectory histories. The dataset further provides 282,952  SAT (Scene--Action association Token) annotations. These annotations link driving scenes, key environmental factors, factor states, and path relations to recorded longitudinal and lateral actions. They support condition-based pairing and association supervision.

Most existing open-loop driving datasets primarily use trajectory errors as evaluation metrics, which do not adequately assess a model's ability to learn scene–action associations. Thus, based on these condition annotations, we propose two metrics, Conditional Response Difference Error (CRDE) and Cross-Environment Conditional Accuracy (CECA). CRDE uses paired samples with different driving conditions within the same environment. It measures the error between predicted and recorded response differences. CECA uses samples with matched driving conditions across different environments. It measures agreement between predicted and recorded longitudinal and lateral actions. The two metrics characterize action adjustments under condition changes and action correctness across environments, respectively. Together with Average Displacement Error (ADE) and Final Displacement Error (FDE), they evaluate planning outputs.

SAT can serve as auxiliary supervision during training to study the role of explicit scene--action information in planning. We analyze changes in trajectory accuracy and conditional responses through baseline comparisons and supervision component ablations. Experimental results show different performance rankings across the two types of metrics. SAT may also reduce trajectory errors while affecting longitudinal and lateral responses differently. These findings show that combining trajectory and conditional response evaluation provides a more specific assessment of how models use scene--action associations.

Our contributions are as follows:
\begin{itemize}
    \item We introduce \textbf{SADrive}, a dataset covering multiple driving environments for scene--action association learning and benchmarking E2E autonomous driving systems. It contains 2,942 driving clips, totaling approximately 13 hours, and covers diverse driving action scenarios.
    \item We establish a benchmark for scene--action association learning. Condition-based pairing, CRDE, and CECA evaluate action adjustments under condition changes and action correctness across environments, respectively.
    \item We provide 282,952 structured SAT scene--action association annotations to study the value of explicit association information for learning to plan.
\end{itemize}

%% file: sec/2_related_work.tex
\section{Related Work}
\subsection{Driving Datasets and Planning Evaluation}
Real-world driving data provide a foundation for studying environment understanding and action decisions. nuScenes~\citep{caesar2020nuscenes} and the Waymo Open Dataset~\citep{sun2020scalability} provide multi-sensor observations and object annotations to support driving scene perception. Argoverse 2~\citep{wilson2023argoverse} and the Waymo Open Motion Dataset~\citep{ettinger2021large} further provide traffic participant trajectories and map information to support motion prediction and interaction modeling. Ithaca365~\citep{diazruiz2022ithaca365} repeatedly collects data across urban, rural, and campus environments, providing data for studying changes in weather, time of day, and traffic conditions. For planning evaluation, nuPlan~\citep{karnchanachari2024nuplan} combines real-world driving records, closed-loop simulation, and planning metrics to evaluate planners. NAVSIM~\citep{dauner2024navsim} uses real observations and short-term non-interactive simulation to evaluate driving quality, including driving progress and collision risk. Bench2Drive~\citep{jia2024bench2drive} organizes interactive scenarios and driving ability tests in CARLA, while RiskBench~\citep{kung2024riskbench} supports risk identification and decision evaluation through scenario classification. For complex real-world scenarios, WOD-E2E~\citep{xu2025wode2e} provides long-tail driving clips and supplements geometric errors with human trajectory preference evaluation, reflecting the need to evaluate action appropriateness.  SADrive further organizes open roads, closed campuses, and transition sections under a unified collection and annotation system. It uses condition-based pairing to examine action adjustments corresponding to changes in driving conditions and action correctness under similar driving conditions across environments, providing data and evaluation support for studying scene--action relationships.

\subsection{Driving Semantics and Scene--Action Associations}
Driving semantics research connects environmental information with driving actions through language descriptions and structured annotations. BDD-X~\citep{kim2018textual} provides descriptions and explanations of driving actions; CoVLA~\citep{arai2025covla} aligns real videos, language descriptions, and trajectories; and LingoQA~\citep{marcu2024lingoqa} evaluates driving scene understanding through video question answering. DriveLM~\citep{sima2024drivelm} and Reason2Drive~\citep{nie2024reason2drive} further organize reasoning relationships among perception, prediction, and planning, providing a data foundation for scene--action association research. Building on this, DriveGPT4~\citep{xu2024drivegpt4} jointly generates action explanations and control signals. VLM-AD~\citep{xu2024vlm} uses reasoning information and action labels generated by vision-language models to assist planning training. Alpamayo-R1~\citep{wang2025alpamayo} integrates decision-related reasoning with trajectory planning. XCoT-VLA~\citep{team2026xcot} organizes trajectory-derived action evidence into scene-grounded tokens for trajectory generation. Recent work further studies the links between environmental dynamics and action decisions, OneWM~\citep{su2026onewm} jointly learns action-conditioned world prediction and action generation for robotic manipulation; SafeDriveVLA~\citep{xie2026safedrivevla} identifies conflicts between instructions and the environment through navigation-conditioned future scene prediction; and AffordDrive3D~\citep{cai2026afforddrive3d} provides action-related spatial information for planning by predicting drivable regions, collision-critical regions, and future geometry. SADrive supports scene--action association research across driving environments from a data and evaluation perspective. It uses condition-based pairing to examine action adjustments corresponding to changes in driving conditions and action correctness across environments, and provides structured SAT annotations to examine the role of scene--action association supervision in learning to plan.

%% file: sec/3_dataset.tex
\section{SADrive Dataset}
\label{sec:dataset}

\begin{figure}[!htbp]
\centering
\IfFileExists{figures/dataset_vis.jpg}{%
  \includegraphics[width=1\textwidth]{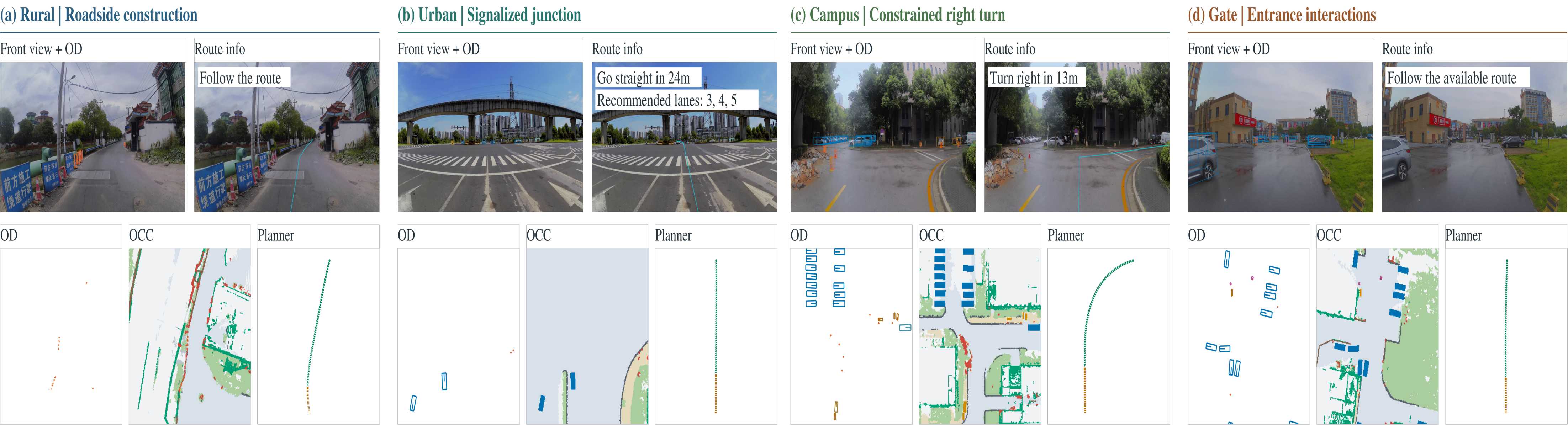}%
}{%
  \missingimage{figures/dataset_vis.jpg}%
}
\caption{\textbf{Representative environments and annotations.} Camera observations, Routes info, 3D object annotations (OD), semantic occupancy annotations(OCC), and ego-vehicle trajectory histories are shown for rural roadside construction, an urban signalized intersection, a campus, and a gate entrance. Spatial annotations provide evidence for analyzing passage factors and path relations, while planning trajectories serve as the main supervision.}
\label{fig:dataset_vis}
\end{figure}

\begin{figure}[!htbp]
\centering
\IfFileExists{figures/data_view.jpg}{%
  \includegraphics[width=0.9\linewidth]{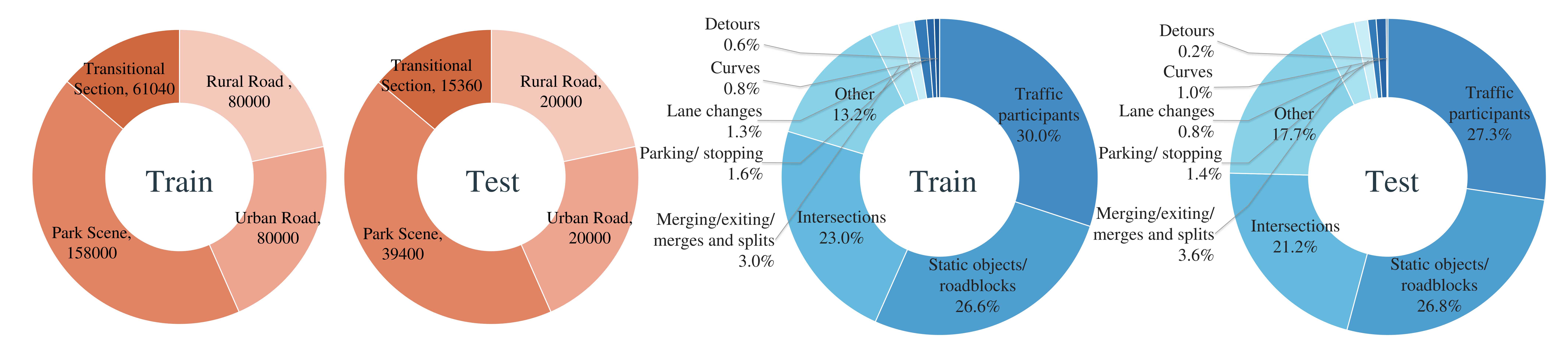}%
}{%
  \missingimage{figures/data_view.jpg}%
}
\caption{\textbf{Training and test data distribution by driving scene.} The distribution summarizes the four environment categories and their frozen split.}
\label{fig:data_view}
\end{figure}

\subsection{Data Distribution}

SADrive covers rural roads, urban roads, campus scenes, and gate-transition sections. Figure~\ref{fig:dataset_vis} shows representative environments and annotations. The dataset contains 2,942 continuous driving clips and 473,800 stored multi-camera frames sampled at 10~Hz. Each stored frame includes object, occupancy, and trajectory labels. SAT annotations are generated for 414,547 frames.
These records undergo automated validation followed by
manual review of the filtered candidates. The final
condition-valid subset contains 282,952 frames with
valid FACTOR, STATE, and REL labels. Another 131,595
records retain action or partial-slot information, but
their condition slots are masked. SADrive are organized by environment, passage factors, and driving actions, covering traffic participants, static obstacles, intersections, parking, lane changes, merges and splits, curves, and detours. Figure~\ref{fig:data_view} summarizes the distribution across these dimensions. SADrive connects diverse road backgrounds with concrete passage conditions, allowing models to adjust actions when key factors change and enabling comparison of planning responses to similar passage problems across environments. The scene, object, occupancy, and trajectory annotations provide a common basis for scene understanding, planning learning, and scene--action evaluation.


\subsection{Sensor Configuration}

\begin{wrapfigure}{l}{0.48\linewidth}
\centering
\IfFileExists{figures/camera_layout.jpg}{%
  \includegraphics[width=\linewidth]{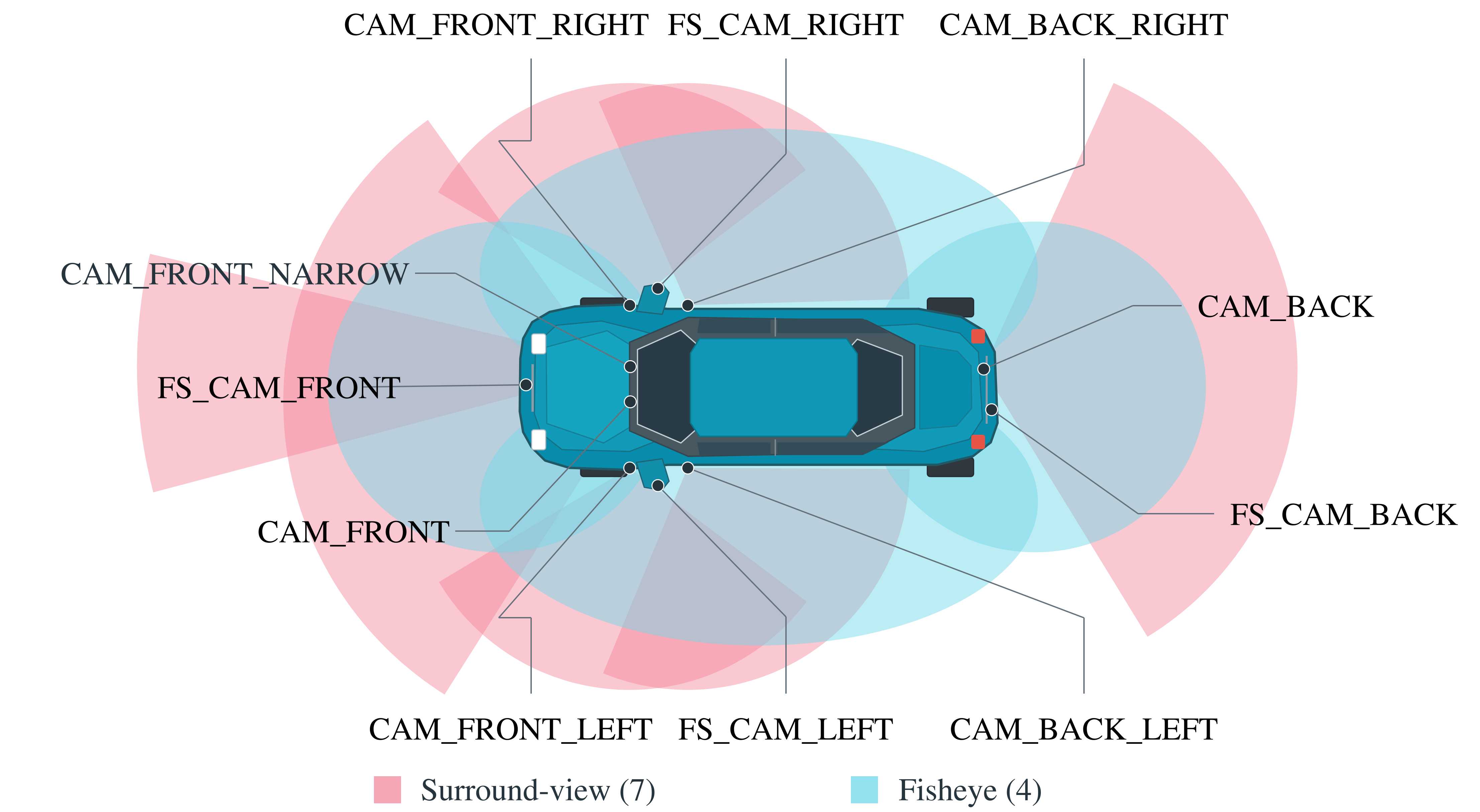}%
}{%
  \missingimage{figures/camera_layout.jpg}%
}
\caption{\textbf{Camera layout.} Seven surround-view and four fisheye channels are labeled with their names. The vehicle faces left, with its right side at the top; positions and coverage are schematic.}
\label{fig:camera_layout}
\end{wrapfigure}As Figure~\ref{fig:camera_layout} shows, the vehicle is equipped with 7 surround-view cameras and 4 fisheye cameras, for a total of 11 channels. The surround-view channels include two front-facing cameras, two on each side, and one rear-facing camera. The fisheye channels face forward, backward, left, and right to supplement near-field observations. The data provide JPEG images and calibration information, including camera intrinsics, distortion parameters, and extrinsics. Planning trajectories are represented as cumulative positions in the ego coordinate system at the anchor, with the data $x$-axis pointing right and the data $y$-axis pointing forward, in meters. The evaluator uses the FLU convention, whose $x$-axis pointing forward and whose y axis points left; all positions and route points are converted by $(x_{\mathrm{FLU}},y_{\mathrm{FLU}})=(y_{\mathrm{data}},-x_{\mathrm{data}})$ before response extraction. The data are organized at 10~Hz, with frame identifiers and timestamps linking images, motion states, and labels. Historical positions are sampled at 20 points from $-2.0$ to $-0.1$ seconds, with the current time set to 0; future positions are sampled at 60 points from $0.1$ to $6.0$ seconds.

\subsection{Planning and Structured Perception Labels}

\begin{figure}[!htbp]
\centering
\IfFileExists{figures/SAT_pipline.jpg}{%
  \includegraphics[width=\linewidth]{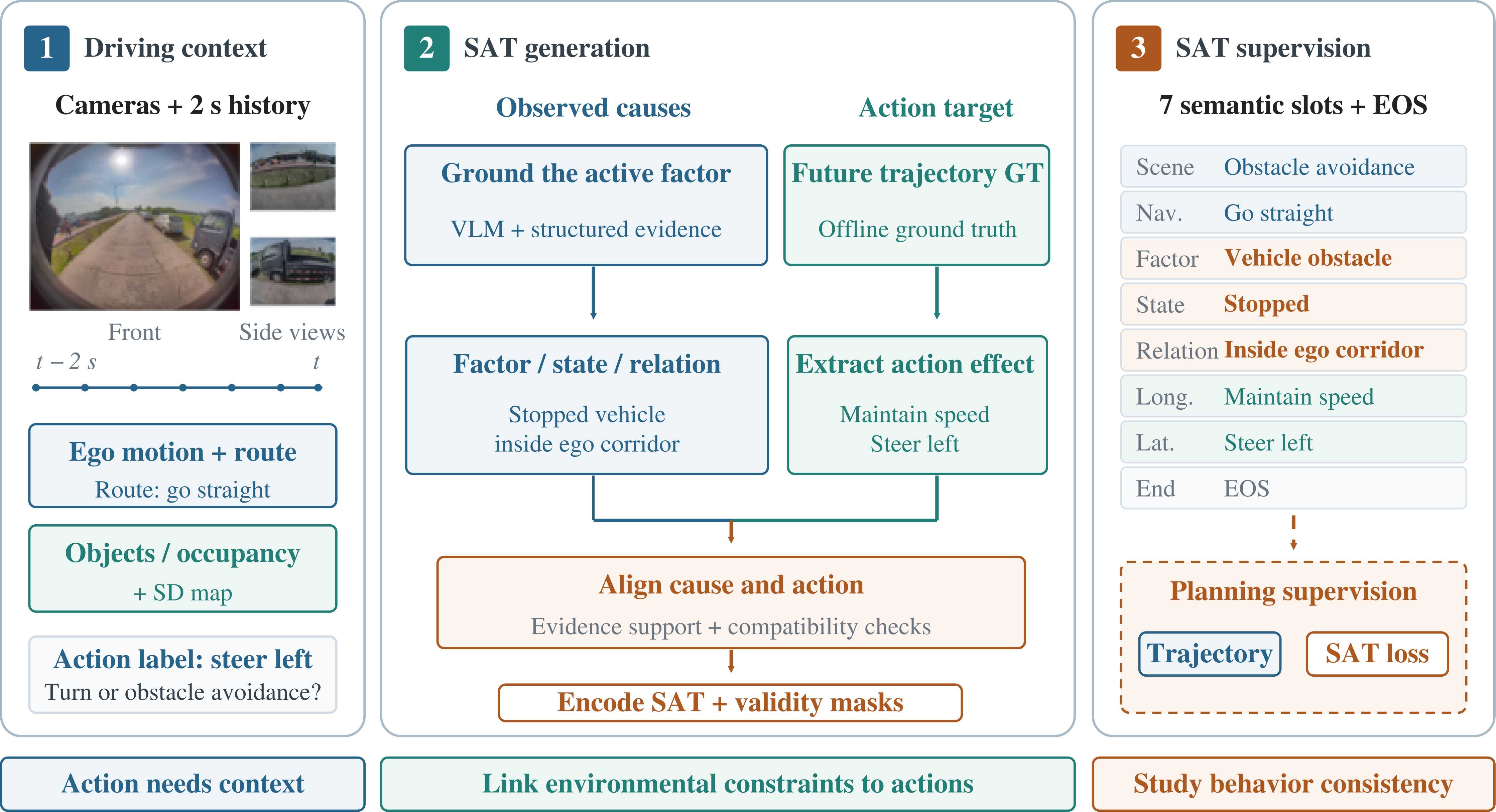}%
}{%
  \missingimage{figures/SAT_pipline.jpg}%
}
\caption{\textbf{Offline SAT annotation pipeline.} Historical and current context provide environmental evidence, while the future recorded trajectory provides a reference for the action. Association information is encoded into a seven-slot sequence and validity mask for auxiliary supervision during training.}
\label{fig:coct_create}
\end{figure}

SADrive provides frame-level information including motion states, future trajectories, surrounding objects, occupancy grids, and available SAT annotations. Ego records provide position, pose, speed, and acceleration, while future trajectories are extracted from continuous driving records. Navigation information comes from local maps and route records, including route polylines, upcoming navigation actions and distances, and road context and recommended lanes when available. Object labels include class, position, size, orientation, future trajectories, and tracking identifiers. Occupancy labels use a local grid with 0.2-meter resolution in both the lateral and longitudinal directions and record semantic and height states. The SD map is used as offline annotation context when available; it is distinct from the route supplied to the planning model. Standard planning inputs consist of historical and current camera observations, ego states, route info and navigation commands. Objects, occupancy, and SAT are available as auxiliary supervision during training, while planning trajectories provide the main planning supervision and evaluation target.
\subsection{Structured SAT Annotations}
\label{sec:coct}
SADrive proposes SAT(Scene--Action association Token) to explicitly represent the links between environmental factors and driving actions. The same action may correspond to different scene situations, so the annotation retains both the action and its related scene information to study the value of scene--action association supervision. As Figure~\ref{fig:coct_create} shows, following a staged annotation process~\citep{team2026xcot}, the construction consists of scene--action pair generation and fixed-slot encoding.

Let $\tau_t$ denote the future recorded trajectory and $e_t$ the ego state at anchor $t$. The auxiliary action label is extracted as
\begin{equation}
A_t^{\mathrm{SAT}}=f_{\mathrm{SAT}}(\tau_t,e_t,\mathbf r_t).
\end{equation}
Here, $\mathbf r_t$ is the route and navigation context, $f_{\mathrm{SAT}}$ is the action extraction function, and $A_t^{\mathrm{SAT}}$ contains the longitudinal and lateral action labels. The condition candidates are generated from current and historical structured observations.

\textbf{Scene--action pair construction.} Longitudinal and lateral actions in the initial action segment are extracted by $f_{\mathrm{SAT}}$ from speed changes, lateral displacement, path shape, and navigation context in the recorded trajectory. The extracted actions cover deceleration, stopping, turning, maintaining speed, maintaining direction, and continued waiting. Candidate factors are constructed from structured current and historical evidence; action compatibility filters whether these candidates can be assigned as the condition triplet. A vision-language model be used the resulting structured evidence at keyframes, while deterministic rules handle the released non-keyframes. The resulting annotations retain both observational provenance and recorded-response compatibility for training supervision.

\textbf{Slot encoding.}
The sequence contains seven slots in order: scene, navigation intent, key factor, factor state, path relation, longitudinal action effect, and lateral action effect, followed by EOS. Factors, states, and relations describe an action-compatible passage condition, while the action slots summarize the recorded response. Fixed slots provide a consistent semantic representation across environments. The annotations retain their source and validity mask for auxiliary supervision during training, while standard planning inputs are used at inference.

%% file: sec/4_benchmark.tex
\section{Unified Planning and Scene--Action Evaluation Benchmark}

SADrive evaluates planning models from two interconnected perspectives, namely trajectory prediction and action generation. Trajectory accuracy measures how close predictions are to driving records; scene--action evaluation examines action adjustments when key factors change and responses to similar passage problems across contexts. SADrive combines ADE/FDE, CRDE, and CECA to evaluate the same model under a unified planning task.
\label{sec:joint_protocol}\label{sec:metrics}
\begin{figure}[!htbp]
\centering \small
\IfFileExists{figures/SADrive_eval.jpg}{%
  \includegraphics[width=1\linewidth]{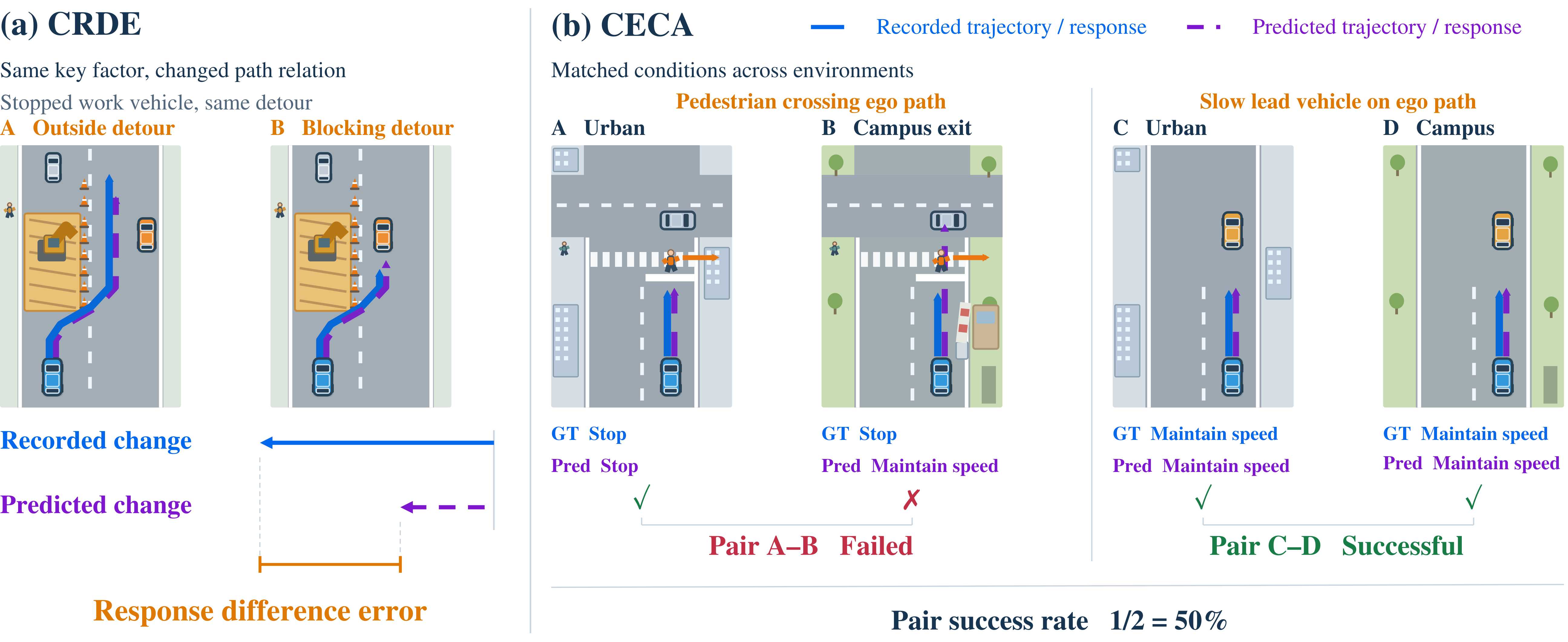}%
}{%
  \missingimage{figures/SADrive_eval.jpg}%
}
\caption{\textbf{Illustration of CRDE and CECA.}
(a) CRDE measures predicted and recorded response differences under changed driving conditions.
(b) CECA evaluates action correctness under matched conditions across environments. 
Blue solid and purple dashed lines show recorded and predicted trajectories, respectively.}
\label{fig:sadrive_eval}
\end{figure}

\subsection{Trajectory Evaluation} 

SADrive report ADE and FDE across four environments, including rural roads (R), urban roads (U), closed campuses (P), and gate transitions (T). For each environment, errors are first averaged over valid anchors within each clip and then averaged equally across clips.
\begin{equation}
\mathrm{ADE}_d =
\frac{1}{|\mathcal C_d|}
\sum_{c\in\mathcal C_d}
\frac{1}{|A_c|}
\sum_{i\in A_c}
\frac{1}{60}\sum_{k=1}^{60}
\|\hat{\mathbf p}_{i,k}-\mathbf p_{i,k}\|_2.
\end{equation}
\begin{equation}
\mathrm{FDE}_d =
\frac{1}{|\mathcal C_d|}
\sum_{c\in\mathcal C_d}
\frac{1}{|A_c|}
\sum_{i\in A_c}
\|\hat{\mathbf p}_{i,60}-\mathbf p_{i,60}\|_2.
\end{equation}
Here, $\mathcal C_d$ denotes the set of valid clips in environment $d$, and $A_c$ represents the set of valid evaluation anchors in clip $c$. The vectors $\hat{\mathbf p}_{i,k}$ and $\mathbf p_{i,k}$ represent the predicted and recorded positions at anchor $i$ and future step $k$, respectively. The 60 prediction steps span 6 seconds at 10~Hz. Macro scores are computed as the average of the four environment-level scores. Both metrics are measured in meters, with lower values indicating better trajectory prediction accuracy.

\subsection{Conditional Response Difference Error}
\label{sec:crde}

CRDE measures the error between predicted and recorded
response differences under changed key-factor conditions,
as illustrated in Figure~\ref{fig:sadrive_eval}(a).
Pairs come from different trips within the same scene.
Navigation intent and the key factor remain fixed.
Either the factor state or the path relation changes,
while the other remains matched. The pairs use the SAT condition-valid subset. We provide an introduction to the Conditional Evaluation Subset in Appendix B.
Responses are extracted from the first 2 seconds of the canonical 10~Hz trajectories. The longitudinal response is the change from current speed
to terminal speed, estimated over the final 0.2 seconds.
The lateral response is the change in signed route-relative offset.
Predictions and records use the same extraction rules of current speed and route geometry.
For direction $\ell\in\{\mathrm{long},\mathrm{lat}\}$,
let $\hat r_i^\ell$ and $r_i^\ell$ denote the predicted
and recorded responses of sample $i$.
The error for paired samples $(i,j)$ is
\begin{equation}
e_{ij}^{\ell}
=
\left|
(\hat r_i^\ell-\hat r_j^\ell)
-
(r_i^\ell-r_j^\ell)
\right|.
\end{equation}
CRDE first averages pair errors within each condition-change
group, then equally averages groups within each environment
and environments.
Longitudinal and lateral CRDE are reported separately
in m/s and m, respectively.
Lower values indicate closer agreement with recorded
response differences.
Absolute endpoint response errors are reported in Appendix E.2.

\subsection{Cross-Environment Conditional Accuracy}
\label{sec:ceca}

CECA evaluates action correctness under matched
key-factor conditions under different environments.
Each pair consists of two samples from different
environments and routes, with matching navigation
intent, key factor, factor state, and path relation.
Each sample retains its own recorded trajectory
and reference action. The evaluation function $f_{\mathrm{eval}}$
independently extracts longitudinal and lateral
action classes from the first 2 seconds of each
trajectory. Longitudinal actions depend on current
and terminal speeds, while lateral actions are
derived from route-relative offsets and route-tangent
changes. The mapping to SAT action labels is provided
in Appendix D.2.

For direction $\ell \in \{\mathrm{long}, \mathrm{lat}\}$, let $\hat{b}_i^\ell$ and $b_i^\ell$ be the predicted and recorded action classes for sample $i$. Endpoint correctness is defined as:
\begin{equation}
a_i^\ell = \mathbb{I}(\hat{b}_i^\ell = b_i^\ell).
\end{equation}
CECA evaluates endpoint accuracy through a hierarchical averaging scheme: correctness is first averaged within each condition group, then equally weighted across groups within each environment, and finally averaged across all environments. Longitudinal and lateral metrics are reported independently.
To capture inter-dependencies, we conduct a pair-wise analysis for matched sample pairs $(i, j)$, evaluating whether both endpoints are correctly predicted:
\begin{equation}
a_{ij}^\ell = a_i^\ell a_j^\ell.
\end{equation}
As illustrated in Figure~\ref{fig:sadrive_eval}(b), a pair is deemed correct if and only if both constituent predictions match their respective references. The pair accuracy is similarly aggregated (first within condition groups, then equally across groups and unordered environment pairs). 

%% file: sec/5_experiments.tex
\providecolor{groupbg}{RGB}{232,236,250}
\providecommand{\pending}{\textemdash}
\providecommand{\cmark}{\ding{51}}
\providecommand{\xmark}{\ding{55}}
\section{Experiments}
\label{sec:experiments}

\subsection{Experimental Settings}
We use 8 NVIDIA H20 GPUs to predict 6-second trajectories from 2-second image/ego histories and navigation. All methods use the same 11-camera image input (7 surround-view and 4 fisheye cameras), ego status, and navigation commands. In this experiment, we select three representative world-model-based approaches. For SADrive, we employ a unified set of evaluation metrics and a consistent protocol for pairing scenario conditions. The dataset is sampled at 2 Hz and comprises 66,376 training anchor boxes and 16,594 testing anchor boxes. The fixed list of test pairs includes 188 CRDE pairs and 236 CECA pairs for each response direction.
\begin{table*}[t]
\centering
\footnotesize
\setlength{\tabcolsep}{2.5pt}
\renewcommand{\arraystretch}{1.05}
\caption{\textbf{Baseline trajectory accuracy and conditional responses.}
ADE/FDE are measured over six seconds in meters;
Macro ADE/FDE weight the four environments equally.
CRDE is reported in m/s longitudinally and m laterally.
CECA reports cross environment conditional accuracy (\%).
All methods use 11 cameras, including 7 surround-view
and 4 fisheye cameras.}
\label{tab:baseline}
\resizebox{\textwidth}{!}{%
\begin{tabular}{l|c|*{14}{r}}
\toprule
\multirow{2}{*}{Method}  
& \multirow{2}{*}{Sensor} 
& \multicolumn{2}{c}{Rural (R)}
& \multicolumn{2}{c}{Urban (U)}
& \multicolumn{2}{c}{Campus (P)}
& \multicolumn{2}{c}{Transition (T)}
& \multicolumn{2}{c}{Macro}
& \multicolumn{2}{c}{CRDE$\downarrow$}
& \multicolumn{2}{c}{CECA$\uparrow$}\\
\cmidrule(lr){3-4}
\cmidrule(lr){5-6}
\cmidrule(lr){7-8}
\cmidrule(lr){9-10}
\cmidrule(lr){11-12}
\cmidrule(lr){13-14}
\cmidrule(lr){15-16}
& &
ADE$\downarrow$ & FDE$\downarrow$ &
ADE$\downarrow$ & FDE$\downarrow$ &
ADE$\downarrow$ & FDE$\downarrow$ &
ADE$\downarrow$ & FDE$\downarrow$ &
ADE$\downarrow$ & FDE$\downarrow$ &
Long. & Lat. & Long. & Lat.\\

\midrule
\rowcolor{groupbg}
\multicolumn{16}{l}{\textit{Implicit world modeling-based}}\\
Drive-JEPA~\cite{wang2026drivejepa} & 11C
& 1.559 & 4.720
& 1.753 & 6.055
& 1.105 & 2.906
& 2.390 & 8.285
& 1.702 & 5.492
& 0.948 & \textbf{0.159}
& 54.48 & 81.76\\

DA-WAM~\cite{zhong2026dawam} & 11C
& 1.444 & 3.881
& 1.214 & 3.372
& 1.096 & 2.917
& 1.317 & 3.580
& 1.268 & 3.437
& \textbf{0.730} & 0.218
& 48.93 & 79.40\\

\midrule
\rowcolor{groupbg}
\multicolumn{16}{l}{\textit{Explicit world generation-based}}\\
DriveLaW~\cite{xia2026drivelaw} & 11C
& \textbf{1.261} & \textbf{3.367}
& \textbf{1.115} & \textbf{3.070}
& \textbf{1.021 }& \textbf{2.708}
& \textbf{1.157} & \textbf{3.157}
& \textbf{1.139} & \textbf{3.075}
& 0.831 & 0.268
& 44.38 & \textbf{84.27}\\

\midrule
\rowcolor{groupbg}
\multicolumn{16}{l}{\textit{World-model-based trajectory evaluation}}\\
WoTE~\cite{li2025endtoenddrivingonlinetrajectory} & 11C
& 1.574 & 4.084
& 1.453 & 3.728
& 1.313 & 3.381
& 2.093 & 5.031
& 1.608 & 4.056
& 0.875 & 0.287
& \textbf{57.90} & 72.70\\
\bottomrule
\end{tabular}
}
\end{table*}

\subsection{Planning across Environments}

DriveLaW achieves the lowest trajectory errors across all four environments, while DA-WAM obtains the lowest longitudinal CRDE and Drive-JEPA achieves the lowest lateral CRDE (Table~\ref{tab:baseline}). WoTE leads longitudinal CECA, whereas DriveLaW achieves the highest lateral CECA. Transition errors are highest for Drive-JEPA and WoTE, while DA-WAM and DriveLaW exhibit their largest errors on rural roads. These differing rankings show that trajectory accuracy does not necessarily reflect conditional response quality, highlighting the complementary value of CRDE and CECA in evaluating scene--action associations.

\begin{table}[!htbp]
\centering\footnotesize
\setlength{\tabcolsep}{2pt}
\renewcommand{\arraystretch}{1.05}
\caption{\textbf{SAT supervision and conditional responses.} CRDE: longitudinal (m/s), lateral (m); CECA: endpoint accuracy (\%). Bold marks the better value within each backbone pair.}
\label{tab:conditional}
\begin{tabularx}{\columnwidth}{@{}l|c|*{6}{>{\raggedleft\arraybackslash}X}@{}}
\toprule
\multirow{2}{*}{Method}  
& \multirow{2}{*}{SAT}
& \multicolumn{2}{c}{Macro (m)} & \multicolumn{2}{c}{CRDE$\downarrow$} & \multicolumn{2}{c}{CECA$\uparrow$}\\
\cmidrule(lr){3-4}\cmidrule(lr){5-6}\cmidrule(lr){7-8}
 & & ADE$\downarrow$ & FDE$\downarrow$ & Long. & Lat. & Long. & Lat.\\

\midrule
Drive-JEPA~\cite{wang2026drivejepa} & \xmark & 1.702 & 5.492
& 0.948 & 0.159
& 54.48 & 81.76 \\
\rowcolor{groupbg}
Drive-JEPA~\cite{wang2026drivejepa} & \cmark & \textbf{1.434} & \textbf{4.501} & \textbf{0.941} & \textbf{0.148} & \textbf{54.58} & \textbf{85.17}\\
\midrule

DA-WAM~\cite{zhong2026dawam} & \xmark & 1.268 & 3.437
& 0.730 & 0.218
& 48.93 & 79.40\\
\rowcolor{groupbg}
DA-WAM~\cite{zhong2026dawam} & \cmark & \textbf{1.267} & \textbf{3.423}
& \textbf{0.726} & \textbf{0.211}
& \textbf{48.95} & \textbf{81.48}\\
\midrule

DriveLaW~\cite{xia2026drivelaw} & \xmark & 1.139 & 3.075 & 0.831 & 0.268 & 44.38 & 84.27\\
\rowcolor{groupbg}
DriveLaW~\cite{xia2026drivelaw} & \cmark & \textbf{1.131} & \textbf{3.036} & \textbf{0.775} & \textbf{0.262} & \textbf{45.78} & \textbf{86.15}\\
\midrule
WoTE~\cite{li2025endtoenddrivingonlinetrajectory} & \xmark & 1.608 & 4.056 & \textbf{0.875} & 0.287 & \textbf{57.90} & 72.70\\
\rowcolor{groupbg}
WoTE~\cite{li2025endtoenddrivingonlinetrajectory} & \cmark & \textbf{1.592} & \textbf{4.016} & 0.896 & \textbf{0.281} & 57.46 & \textbf{74.72}\\
\bottomrule
\end{tabularx}
\end{table}

\begin{figure}[!htbp]
\centering \small
\IfFileExists{figures/metric_validity.jpg}{%
\includegraphics[width=1\linewidth]{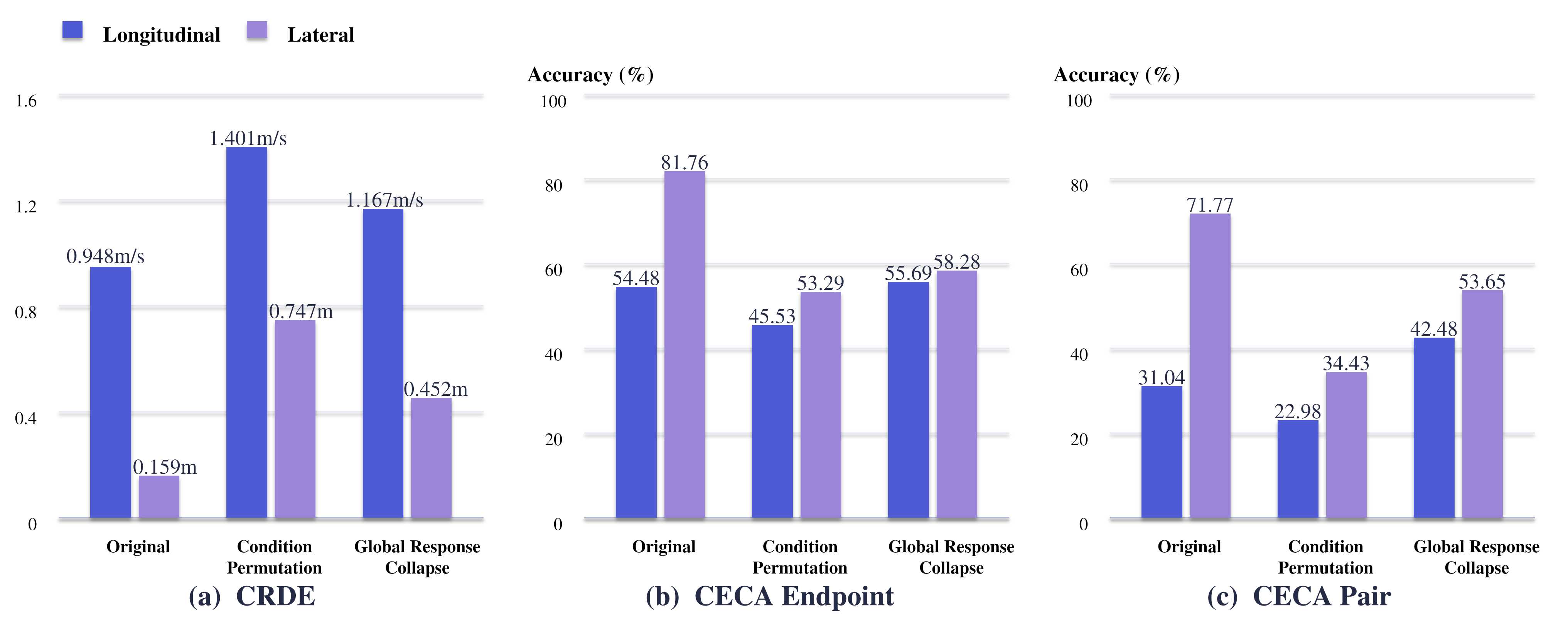}%
}{
  \missingimage{figures/metric_validity.jpg}%
}
\caption{\textbf{Metric validity under condition-insensitive
    controls.}
    (a) CRDE measures response-difference agreement.
    (b) CECA endpoint accuracy evaluates individual action
    correctness.
    (c) CECA pair accuracy evaluates joint correctness across
    matched environments.
    Results are shown for original Drive-JEPA predictions,
    condition permutation, and global response collapse.
    Lower CRDE and higher CECA indicate better performance.}
\label{fig:metric_validity}
\end{figure}

\subsection{Scene--Action Supervision}

In Table~\ref{tab:conditional}, SAT reduces macro ADE/FDE across all backbones, but its effects on conditional responses vary. All backbones improve in lateral CRDE and CECA, whereas WoTE deteriorates longitudinally despite lower trajectory errors. This divergence demonstrates that CRDE and CECA capture scene--action response differences beyond trajectory accuracy.

\begin{figure}[!t]
\centering \small
\IfFileExists{figures/eval_vis.jpg}{%
  \includegraphics[width=\linewidth]{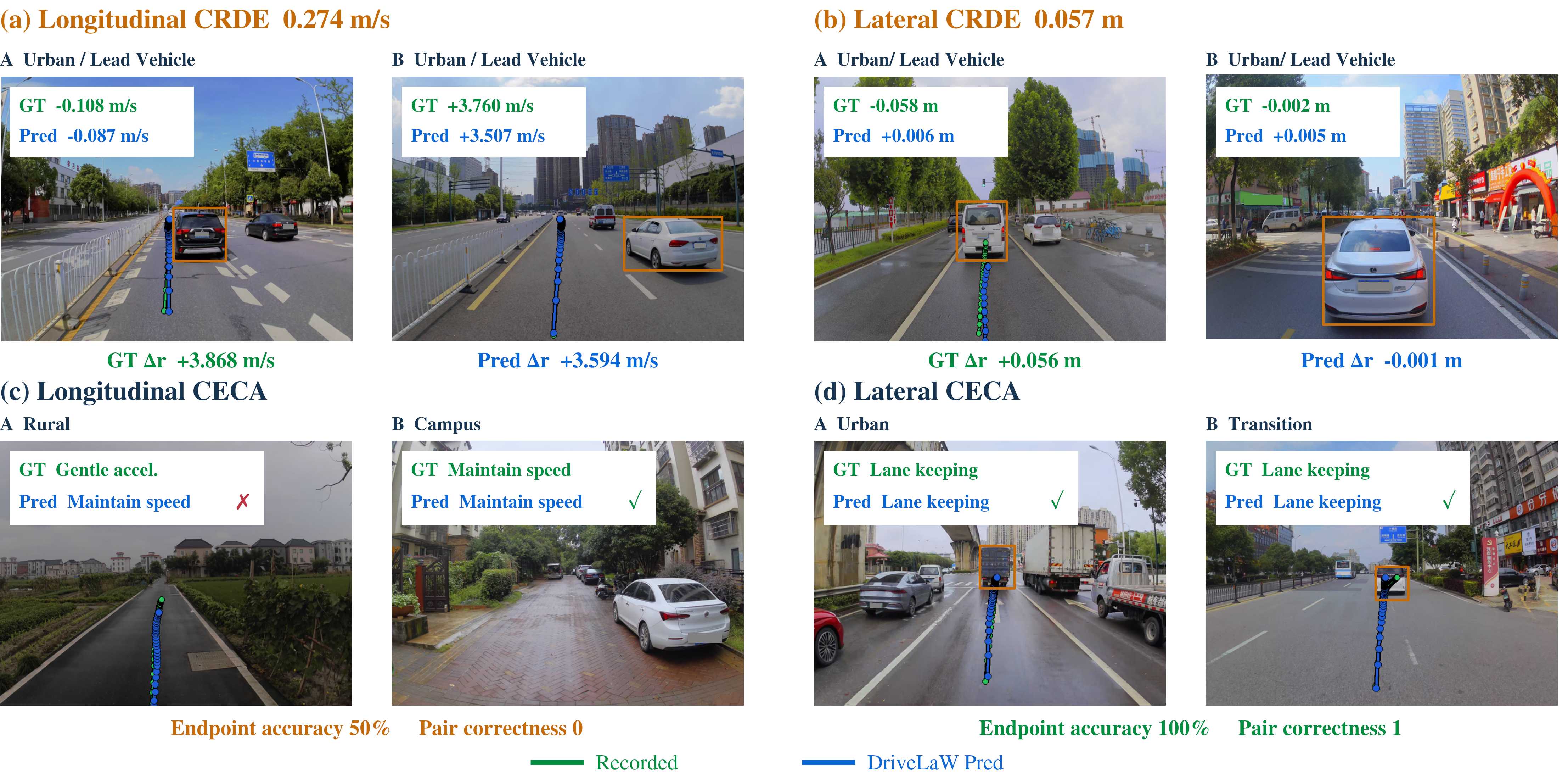}%
}{%
  \missingimage{figures/eval_vis.jpg}%
}
\caption{\textbf{Qualitative analysis of CRDE and CECA.}
(a--b) CRDE evaluates longitudinal and lateral response
differences when the same key factor changes state or
path relation within an environment.
(c--d) CECA evaluates longitudinal and lateral action
correctness under matched key-factor conditions across environments.
Displayed accuracies and pair scores refer to the example pairs.
Green and blue denote recorded and predicted trajectories,
respectively.}
\label{fig:eval_vis}
\end{figure}

\subsection{Metric Validity Analysis}

We further examine whether the proposed metrics can identify failures in condition-dependent action prediction. To this end, we construct two condition-insensitive controls using Drive-JEPA trained without SAT supervision. Condition-response permutation disrupts the correspondence between scene conditions and predicted responses, while global response collapse removes sample-specific response variations.

As shown in Figure~\ref{fig:metric_validity}, condition permutation increases CRDE from 0.948 to 1.401 longitudinally and from 0.159 to 0.747 laterally. It also reduces CECA endpoint accuracy from 54.48\% to 45.53\% and from 81.76\% to 53.29\%, respectively. Global response collapse increases CRDE in both directions and reduces lateral CECA, yet slightly improves aggregate longitudinal CECA. 

\begin{wraptable}{l}{0.48\linewidth}
\centering
\footnotesize
\setlength{\tabcolsep}{3pt}
\renewcommand{\arraystretch}{1.1}
\caption{\textbf{CECA pair accuracy under different reference-action relationships.} Pairs are grouped using reference action labels.}
\label{tab:ceca_action_groups}
\resizebox{\linewidth}{!}{%
\begin{tabular}{llrr}
\toprule
Setting & Dimension & Same-action & Different-action\\
\midrule
\multirow{2}{*}{Original}
& Long. & 40.87 & 20.71\\
& Lat. & 81.47 & 53.68\\
\midrule
\multirow{2}{*}{Permutation}
& Long. & 34.41 & 11.16\\
& Lat. & 47.23 & 12.69\\
\midrule
\multirow{2}{*}{Global Collapse}
& Long. & 81.38 & 1.26\\
& Lat. & 85.92 & 0.00\\
\bottomrule
\end{tabular}%
}
\end{wraptable}The pair-wise results in Table~\ref{tab:ceca_action_groups} explain this discrepancy. Global response collapse achieves 81.38\% and 85.92\% on Same-action pairs, but drops to 1.26\% and 0.00\% on Different-action pairs. 
In contrast, original Drive-JEPA achieves 20.71\% and 53.68\% on Different-action pairs, while condition permutation reduces them to 11.16\% and 12.69\%. These results confirm that aggregate accuracy can conceal failures under different action requirements, highlighting the value of pair-wise CECA.


\subsection{Qualitative Analysis}
\label{sec:qualitative}

Figure~\ref{fig:eval_vis} analyzes scene--action associations
through paired DriveLaW predictions.
In (a), the model predicts near-constant speed in A and
acceleration in B.
This agrees with the recorded longitudinal response change,
yielding a CRDE of 0.274\,m/s.
In (b), the predicted lateral response difference is nearly
zero ($-0.001$\,m).
The recorded difference is also small ($+0.056$\,m),
yielding a CRDE of 0.057\,m.
Thus, CRDE requires interpretation alongside the recorded
response scale.
In (c), speed maintenance matches the campus reference
but misses rural acceleration, yielding 50\% endpoint
accuracy and pair correctness of 0.
In (d), both lane-keeping predictions match their references,
yielding 100\% endpoint accuracy and pair correctness of 1.
These examples illustrate how CECA evaluates whether
predictions under similar conditions match the recorded
actions in each environment.

%% file: sec/6_conclusion.tex
\section{Discussion and Conclusion}

We present \textbf{SADrive}, a real-world dataset for studying scene--action associations in end-to-end autonomous driving. Beyond trajectory prediction, \textbf{SADrive} examines how driving actions respond to changing environmental conditions and remain consistent across environments. To this end, we introduce CRDE and CECA as complementary metrics for evaluating condition-dependent action responses. Experiments show that improved trajectory accuracy does not necessarily lead to better conditional responses. Moreover, the effects of SAT supervision vary across planning backbones and action directions. These findings highlight the need to evaluate trajectory accuracy and scene--action associations jointly. \textbf{SADrive} provides a basis for assessing condition-dependent planning and understanding how scene information shapes driving actions.

%% file: sec/LLM_Usage_Statement.tex
\section*{LLM Usage Statement}

We used OpenAI GPT to assist with generating SAT annotations, improving the language and clarity of the manuscript, and providing suggestions for figure layouts and visual design. The authors take full responsibility for the dataset annotations, scientific claims, experimental results, and final manuscript.

%% file: sec/X_suppl.tex
\providecommand{\pending}{\textemdash}
\providecommand{\cmark}{\ding{51}}
\providecommand{\xmark}{\ding{55}}
\appendix

\section{SAT Annotation and Auxiliary Supervision}
\label{sec:coct_vocab_semantics}

\subsection{Condition Grounding and Slot Validity}
\label{app:compatibility}

The SAT scene slot describes a local interaction, such as following or obstacle avoidance. It is distinct from the acquisition environments used for trajectory evaluation. Navigation intent describes the route task. The key factor, factor state, and path relation specify the passage condition. Longitudinal and lateral action slots describe the recorded response. Table~\ref{tab:coct_vocab} lists the complete token classes for all seven slots. Tokens containing NONE indicate the absence of a valid factor or action, UNOBSERVED indicates insufficient observation support, and UNKNOWN indicates that a more specific class cannot be assigned.

\begin{table}[!htbp]
\centering\scriptsize
\setlength{\tabcolsep}{3pt}
\renewcommand{\arraystretch}{1.12}
\caption{\textbf{SAT token vocabulary.} The seven slots are listed in sequence order. $N$ is the number of token classes in each slot. EOS terminates the sequence and is excluded from the semantic-slot classification loss.}
\label{tab:coct_vocab}
\label{tab:condition_fields}
\begin{tabularx}{\textwidth}{@{}>{\raggedright\arraybackslash}p{0.14\textwidth}r>{\raggedright\arraybackslash}X@{}}
\toprule
Slot & $N$ & Token classes\\
\midrule
Scene & 13 & \texttt{SCENE\_SIGNALIZED\_INTERSECTION}, \texttt{SCENE\_INTERSECTION}, \texttt{SCENE\_LANE\_CHANGE}, \texttt{SCENE\_MERGE\_SPLIT}, \texttt{SCENE\_GATE}, \texttt{SCENE\_CROSSING\_VRU}, \texttt{SCENE\_FOLLOWING}, \texttt{SCENE\_OBSTACLE\_AVOIDANCE}, \texttt{SCENE\_STOP\_OR\_PARK}, \texttt{SCENE\_FREE\_SPACE\_MANEUVER}, \texttt{SCENE\_CRUISE}, \texttt{SCENE\_START\_MOVING}, \texttt{SCENE\_UNKNOWN} \\
Navigation intent & 19 & \texttt{NAV\_FOLLOW\_ROUTE}, \texttt{NAV\_GO\_STRAIGHT}, \texttt{NAV\_TURN\_LEFT}, \texttt{NAV\_TURN\_RIGHT}, \texttt{NAV\_U\_TURN}, \texttt{NAV\_MERGE}, \texttt{NAV\_SPLIT}, \texttt{NAV\_EXIT\_LEFT}, \texttt{NAV\_EXIT\_RIGHT}, \texttt{NAV\_PASS\_GATE}, \texttt{NAV\_LANE\_CHANGE\_LEFT}, \texttt{NAV\_LANE\_CHANGE\_RIGHT}, \texttt{NAV\_OVERTAKE}, \texttt{NAV\_FOLLOW\_LEAD}, \texttt{NAV\_YIELD}, \texttt{NAV\_REACH\_STOP\_AREA}, \texttt{NAV\_FREE\_SPACE\_TRAVERSE}, \texttt{NAV\_PARK}, \texttt{NAV\_UNKNOWN} \\
\midrule
Key factor & 22 & \texttt{none}, \texttt{unobserved}, \texttt{route\_requirement}, \texttt{traffic\_light}, \texttt{stop\_line}, \texttt{yield\_sign}, \texttt{gate}, \texttt{gate\_arm}, \texttt{temp\_traffic\_control}, \texttt{traffic\_controller}, \texttt{lead\_vehicle}, \texttt{pedestrian}, \texttt{cyclist}, \texttt{animal}, \texttt{crossing\_vehicle}, \texttt{mixed\_agent}, \texttt{merging\_vehicle}, \texttt{target\_lane\_agent}, \texttt{static\_obstacle}, \texttt{vehicle\_obstacle}, \texttt{road\_geometry}, \texttt{unknown} \\
Factor state & 31 & \texttt{STATE\_NONE}, \texttt{STATE\_UNOBSERVED}, \texttt{STATE\_ACTIVE}, \texttt{STATE\_RED}, \texttt{STATE\_GREEN}, \texttt{STATE\_CLOSED}, \texttt{STATE\_CLOSING}, \texttt{STATE\_LOWERED}, \texttt{STATE\_OPENING}, \texttt{STATE\_OPEN}, \texttt{STATE\_DECELERATING}, \texttt{STATE\_STOPPED}, \texttt{STATE\_SLOW\_MOVING}, \texttt{STATE\_APPROACHING}, \texttt{STATE\_ENTERING\_PATH}, \texttt{STATE\_CROSSING}, \texttt{STATE\_OCCUPYING\_PATH}, \texttt{STATE\_STARTING}, \texttt{STATE\_CLEARING}, \texttt{STATE\_CLEARED}, \texttt{STATE\_GAP\_INSUFFICIENT}, \texttt{STATE\_GAP\_AVAILABLE}, \texttt{STATE\_BLOCKING}, \texttt{STATE\_CURVING\_LEFT}, \texttt{STATE\_CURVING\_RIGHT}, \texttt{STATE\_NARROWING}, \texttt{STATE\_TURN\_REQUIRED}, \texttt{STATE\_LANE\_CHANGE\_REQUIRED}, \texttt{STATE\_MERGE\_REQUIRED}, \texttt{STATE\_PROGRESS\_REQUIRED}, \texttt{STATE\_UNKNOWN} \\
Path relation & 15 & \texttt{REL\_NONE}, \texttt{REL\_UNOBSERVED}, \texttt{REL\_ON\_REF\_PATH}, \texttt{REL\_ON\_ROUTE\_CONTROL}, \texttt{REL\_AHEAD\_SAME\_PATH}, \texttt{REL\_INSIDE\_EGO\_CORRIDOR}, \texttt{REL\_APPROACHING\_EGO\_CORRIDOR}, \texttt{REL\_CROSSING\_EGO\_CORRIDOR}, \texttt{REL\_TARGET\_LANE\_GAP}, \texttt{REL\_BLOCKING\_EGO\_PATH}, \texttt{REL\_BLOCKING\_EGO\_CORRIDOR}, \texttt{REL\_CONTROLLING\_EGO\_CORRIDOR}, \texttt{REL\_NOT\_BLOCKING\_EGO\_CORRIDOR}, \texttt{REL\_OUTSIDE\_EGO\_CORRIDOR}, \texttt{REL\_UNKNOWN} \\
\midrule
Longitudinal action & 7 & \texttt{EFFECT\_LONG\_MAINTAIN}, \texttt{EFFECT\_LONG\_ACCELERATE}, \texttt{EFFECT\_LONG\_DECELERATE}, \texttt{EFFECT\_LONG\_STOP}, \texttt{EFFECT\_LONG\_HOLD}, \texttt{EFFECT\_LONG\_NONE}, \texttt{EFFECT\_LONG\_UNKNOWN} \\
Lateral action & 17 & \texttt{EFFECT\_LAT\_KEEP}, \texttt{EFFECT\_LAT\_TURN\_LEFT}, \texttt{EFFECT\_LAT\_TURN\_RIGHT}, \texttt{EFFECT\_LAT\_U\_TURN}, \texttt{EFFECT\_LAT\_MERGE\_SPLIT}, \texttt{EFFECT\_LAT\_LANE\_CHANGE\_LEFT}, \texttt{EFFECT\_LAT\_LANE\_CHANGE\_RIGHT}, \texttt{EFFECT\_LAT\_NUDGE\_LEFT}, \texttt{EFFECT\_LAT\_NUDGE\_RIGHT}, \texttt{EFFECT\_LAT\_STEER\_LEFT}, \texttt{EFFECT\_LAT\_STEER\_RIGHT}, \texttt{EFFECT\_LAT\_FREE\_SPACE\_STEER\_LEFT}, \texttt{EFFECT\_LAT\_FREE\_SPACE\_STEER\_RIGHT}, \texttt{EFFECT\_LAT\_PULL\_OVER}, \texttt{EFFECT\_LAT\_ABORT}, \texttt{EFFECT\_LAT\_NONE}, \texttt{EFFECT\_LAT\_UNKNOWN} \\
\bottomrule
\end{tabularx}
\end{table}

Candidate conditions are grounded in current and historical observations. Recorded future trajectories provide action labels and a compatibility check. For example, a lead vehicle supports a following response when it is ahead on the same path. A gate-opening event requires evidence of an opening state and a preceding closed or closing state. An open gate alone is context rather than evidence of an opening event. Conditions without sufficient observation support or action compatibility have their condition slots masked. Their remaining valid slots can still contribute supervision. The validity mask therefore combines observational support with recorded-action compatibility.

\subsection{SAT Quality Control}

The SAT annotation pipeline applies automated validation
followed by manual review of the filtered candidates.
Automated validation checks slot formats, label validity,
annotation completeness, cross-slot logical consistency,
and action compatibility. These checks remove records
that violate the predefined annotation rules.

Manual review examines the 11 camera views, 3D annotations,
dynamic object trajectories, and ego motion. Reviewers
check five aspects of the annotations: scene description,
key-factor identification, factor state, path relation,
and ego longitudinal and lateral actions. A key factor
is retained only when the observations support its
potential influence on ego driving. Its state and path
relation are then checked against the visual and
geometric evidence. The recorded ego actions are also
examined for consistency with the identified driving
condition.

The generation process produces 414,547 frame-level
SAT records. After the two-stage quality-control
process, 282,952 records are retained with valid
FACTOR, STATE, and REL labels. Another 131,595 records
retain action or partial-slot information, but their
condition slots are masked. The remaining 59,253
dataset frames have no SAT records. Partially valid
records are preserved for auxiliary supervision rather
than discarded entirely. Only condition-valid records
are considered for condition-based pairing.

\subsection{Masked Auxiliary Objective}
\label{sec:coct_objective}

The auxiliary heads predict the seven SAT slots from the shared planning representation. For training sample $i$ and slot $j\in\{1,\ldots,7\}$, let $\mathbf z_i$ be this representation, $h_j$ the slot classifier, $y_{i,j}$ the target class, and $m_{i,j}\in\{0,1\}$ its validity mask. The training objective is
\begin{equation}
\mathcal L=\mathcal L_{\mathrm{traj}}
+\lambda\frac{\sum_{i,j}m_{i,j}\operatorname{CE}(h_j(\mathbf z_i),y_{i,j})}
{\max(1,\sum_{i,j}m_{i,j})},
\end{equation}
where $\mathcal L_{\mathrm{traj}}$ is the original trajectory objective, $\operatorname{CE}$ denotes cross-entropy, and $\lambda\geq0$ weights auxiliary supervision. The sums run over the training samples in a batch and the seven slots. EOS is a sequence terminator and is excluded from the loss. Samples without valid auxiliary labels retain trajectory supervision. Ground-truth SAT is not supplied at inference.

\section{Conditional Evaluation Subset}
\label{app:conditional_protocol}

\subsection{Eligibility and Pair Construction}

The conditional subset is selected from the 16,594 valid test anchors using the SAT condition annotations. Training and test samples are separated by source trip and event, including overlapping observation and target intervals. Condition pairing is performed offline and fixed before model comparison.

Table~\ref{tab:pairing_rules} summarizes the two pairing rules. CRDE holds scene, navigation intent and key factor fixed while changing exactly one of STATE or REL. The unchanged field remains matched. CECA matches all four condition fields across different environments. Reference action classes are excluded from matching keys, and the two CECA endpoints need not share the same reference action. 

\begin{table}[!htbp]
\centering\small
\setlength{\tabcolsep}{3pt}
\caption{\textbf{Pair construction for conditional evaluation.} Each pair uses samples from different source trips.}
\label{tab:pairing_rules}
\begin{tabularx}{\columnwidth}{@{}l>{\raggedright\arraybackslash}p{0.21\columnwidth}>{\raggedright\arraybackslash}X@{}}
\toprule
Metric & Environment & Condition fields\\
\midrule
CRDE & Same & Match NAV and FACTOR; change exactly one of STATE or REL\\
CECA & Different & Match NAV, FACTOR, STATE, and REL\\
\bottomrule
\end{tabularx}
\end{table}

Both protocols also match current speed, longitudinal acceleration, historical lateral speed, route-relative offset, and route curvature. Features are standardized using training-set scales. Each standardized feature gap is at most 2, and the Euclidean matching distance is at most 3. Factor distance is recorded when available but is not a matching feature. Matching therefore controls the declared condition and motion variables without making the full scenes identical.

Each endpoint is used at most once within a metric and direction. Each source group, identified by the grouping key in the pairing manifest, contributes at most eight endpoints. Scores are aggregated over the supported condition groups and environments in the resulting subset. No quota requires every environment to contribute a valid group. All compared models and supervision settings use the same frozen pairs for each metric and direction.

\subsection{Aggregation and Pair Support}

CRDE averages response-difference errors within each condition-change group, then equally averages group-level scores within each supported environment and environment-level scores across supported
environments. Primary CECA averages endpoint correctness within each condition group and environment, then equally across groups and environments. The companion CECA$_{\mathrm{pair}}$ requires both endpoints to match their own reference actions. It is averaged within each condition group and unordered environment pair, then equally across groups and environment pairs. Longitudinal and lateral scores are computed separately. These conditional averages use the selected subset rather than the four-environment denominator used for trajectory macro scores.

Table~\ref{tab:conditional_support} gives the available support counts  run with SAT. CRDE endpoint counts are taken from the corresponding absolute-response export. The two directions are counted separately because their endpoint sets may overlap.

\begin{table}[!htbp]
\centering\footnotesize
\setlength{\tabcolsep}{3pt}
\caption{\textbf{Reported support per direction with SAT.}}
\label{tab:conditional_support}
\begin{fittable}
\begin{tabular*}{\columnwidth}{@{\extracolsep{\fill}}lrr@{}}
\toprule
Metric & Pairs & Endpoints\\
\midrule
CRDE & 188 & 376\\
CECA & 236 & 472\\
\bottomrule
\end{tabular*}
\end{fittable}
\end{table}

\section{Metric Validity under Condition-insensitive Controls}
\label{app:metric_validity}

\subsection{Evaluation Protocol}
\label{app:metric_validity_protocol}

We conduct metric validity experiments using the Drive-JEPA. Both negative controls operate on the continuous responses extracted from the predicted trajectories. All modified responses are evaluated using the original CRDE and CECA evaluation pipeline. No learnable weights are introduced in either control. Training-set statistics are used only to normalize
ego-motion features for permutation grouping and
to estimate the global response median. All metrics
retain the original macro aggregation rules.

\subsection{Condition--Response Permutation}
\label{app:condition_permutation}

We perform response permutation within groups of
similar ego-motion states. Each test anchor is assigned
to a group according to its acquisition environment,
current speed, longitudinal acceleration, and
historical lateral velocity.

The three motion features are standardized using
the means and standard deviations estimated from
the training cache. For anchor $i$, let
\begin{equation}
z_i^q =
\frac{x_i^q-\mu_q^{\mathrm{train}}}
{\sigma_q^{\mathrm{train}}},
\quad
q\in\{\mathrm{speed},\mathrm{accel},\mathrm{latvel}\}.
\end{equation}
The grouping key is defined as
\begin{equation}
\begin{aligned}
\mathcal G_i = (&e_i,
\lfloor z_i^{\mathrm{speed}}/0.5\rfloor,\\
&\lfloor z_i^{\mathrm{accel}}/0.5\rfloor,
\lfloor z_i^{\mathrm{latvel}}/0.5\rfloor),
\end{aligned}
\end{equation}
where $e_i\in\{R,U,P,T\}$ denotes the acquisition
environment. Each motion feature is thus divided
into bins of 0.5 standardized units.

Let $\mathcal G_g$ contain the anchors assigned
to group $g$, and let
$\hat{\mathbf r}_i =
(\Delta v_i,\Delta l_i,\Delta\psi_i)$ denote the
continuous predicted response of anchor $i$.
For each eligible group, we apply a random non-zero
cyclic shift. If the group contains $n_g$ anchors
ordered as $i_{g,0},\ldots,i_{g,n_g-1}$, a shift
$s_g\in\{1,\ldots,n_g-1\}$ gives
\begin{equation}
\hat{\mathbf r}^{\mathrm{perm}}_{i_{g,j}}
=
\hat{\mathbf r}_{i_{g,(j+s_g)\bmod n_g}}.
\end{equation}
This derangement ensures that no anchor in an
eligible group retains its own response. The three
response dimensions are permuted jointly, preserving
their original within-group joint distribution.

The test set contains 16,650 anchors and 1,915
motion groups. Among them, 1,185 groups are eligible
for permutation, covering 15,920 anchors. The
remaining 730 anchors belong to singleton or
small groups and retain their original responses.

For each permuted anchor, its own observed initial
speed $v_0$ is retained. The terminal speed is
reconstructed as
\begin{equation}
v_2^{\mathrm{perm}}
=
v_0+\Delta v^{\mathrm{perm}}.
\end{equation}
The modified continuous responses are then passed
through the original action classifier to recompute
the evaluation action labels. The reference actions
and all other evaluation targets remain unchanged.

We repeat the procedure 1,000 times using random
seed 20261104. Each repetition uses the same frozen
Pair Manifest and the same eligible anchor set.

\begin{table}[!htbp]
\centering
\footnotesize
\setlength{\tabcolsep}{5pt}
\renewcommand{\arraystretch}{1.1}
\caption{\textbf{Support of condition--response permutation.}}
\label{tab:permutation_support}
\begin{fittable}
\begin{tabular}{lr}
\toprule
Item & Count\\
\midrule
Test anchors & 16,594\\
Motion groups & 1,915\\
Eligible groups & 1,185\\
Permuted anchors & 15,920\\
Unpermuted anchors & 730\\
Permutation repetitions & 1,000\\
Random seed & 20261104\\
\bottomrule
\end{tabular}
\end{fittable}
\end{table}

\begin{table}[!htbp]
\centering
\footnotesize
\setlength{\tabcolsep}{5pt}
\renewcommand{\arraystretch}{1.1}
\caption{\textbf{Statistical results of condition response
permutation.} Results are obtained from 1,000 permutations.
The intervals denote the 2.5th and 97.5th percentiles.}
\label{tab:permutation_statistics}
\begin{fittable}
\begin{tabular}{lrrrrr}
\toprule
Metric & Original & Perm. Mean & Std.
& 95\% Interval & $p$-value\\
\midrule
CRDE Long.
& 0.9479 & 1.4005 & 0.0650
& [1.2768, 1.5340] & 0.001\\
CRDE Lat.
& 0.1586 & 0.7466 & 0.0594
& [0.6469, 0.8610] & 0.001\\
CECA Long. Endpoint
& 54.48 & 45.53 & 1.04
& [43.44, 47.69] & 0.001\\
CECA Long. Pair
& 31.04 & 22.98 & 1.37
& [20.52, 25.63] & 0.001\\
CECA Lat. Endpoint
& 81.76 & 53.29 & 1.13
& [51.04, 55.50] & 0.001\\
CECA Lat. Pair
& 71.77 & 34.43 & 1.55
& [31.61, 37.55] & 0.001\\
\bottomrule
\end{tabular}
\end{fittable}
\end{table}

The empirical $p$-value is computed as
\begin{equation}
p =
\frac{1+N_{\mathrm{extreme}}}
{1+N_{\mathrm{perm}}},
\end{equation}
where $N_{\mathrm{extreme}}$ counts permutations
with results at least as favorable as the original
prediction. Lower CRDE and higher CECA are considered
favorable. All six metrics achieve the minimum
empirical $p$-value of $1/1001\approx0.001$.

\subsection{Global Response Collapse}
\label{app:global_response_collapse}

Global response collapse removes sample-specific
response variations by assigning a shared response
to all test anchors. The fixed response is estimated
using the component-wise median of valid responses
from the training cache. Specifically,
\begin{equation}
\begin{aligned}
\Delta v^{\mathrm{global}} &= 0.006629~\mathrm{m/s},\\
\Delta l^{\mathrm{global}} &= 0.000510~\mathrm{m},\\
\Delta\psi^{\mathrm{global}} &= 0~\mathrm{rad}.
\end{aligned}
\end{equation}

These values are estimated from 66,376 valid
training responses. 

For each test anchor $i$, its observed initial
speed $v_{0,i}$ is retained, while the three
response components are replaced by their
global medians:
\begin{equation}
\hat{\mathbf r}^{\mathrm{collapse}}_i
=
(\Delta v^{\mathrm{global}},
\Delta l^{\mathrm{global}},
\Delta\psi^{\mathrm{global}}).
\end{equation}
The terminal speed is reconstructed as
\begin{equation}
v_{2,i}^{\mathrm{collapse}}
=
v_{0,i}+\Delta v^{\mathrm{global}}.
\end{equation}
The resulting responses are passed through the
same action classifier and CRDE/CECA evaluator.

All 16,594 test anchors produce valid collapsed
responses, with no invalid samples. This control
removes sample-specific response variations while
preserving each anchor's observed initial speed
and the original evaluation protocol.

\begin{table}[!htbp]
\centering
\footnotesize
\setlength{\tabcolsep}{5pt}
\renewcommand{\arraystretch}{1.1}
\caption{\textbf{Training statistics and test support
for global response collapse.}}
\label{tab:collapse_support}
\begin{fittable}
\begin{tabular}{lr}
\toprule
Item & Value\\
\midrule
Training response samples & 66,376\\
Training valid responses & 66,376\\
Training invalid responses & 0\\
Test anchors & 16,594\\
Valid collapsed responses & 16,594\\
Invalid collapsed responses & 0\\
\bottomrule
\end{tabular}
\end{fittable}
\end{table}

\subsection{Reference-Action Pair Stratification}
\label{app:reference_action_stratification}

We further stratify CECA pairs according to their
ground-truth reference action labels. A pair is
classified as Same-action when both endpoints
have identical reference action classes. Otherwise,
it is classified as Different-action. 
For each pair, both predicted actions must match
their respective reference labels to count as correct.
Pair accuracy follows the original CECA macro
aggregation. Scores are first averaged within
condition groups, then equally across groups and
unordered environment pairs.

\section{Response Extraction and Action Labels}

\subsection{Continuous Responses}

Each model prediction contains 12 future
trajectory points at 0.5-second intervals in the
FLU coordinate frame. Following the evaluation pipeline, the anchor state at time zero
is prepended, and the trajectory is linearly
interpolated to 60 future points at 0.1-second
intervals. Responses are then extracted from the
first 2 seconds using the existing response
extraction procedure.
Let $\mathbf p(t)$ be the two-dimensional position in the FLU frame at time $t$ in seconds, and let $v_0$ be the recorded speed at the anchor. Let $\rho(t)$ denote the signed lateral offset of $\mathbf p(t)$ from the current route. The terminal speed $v_2$ and the two responses are
\begin{equation}
\begin{aligned}
v_2&=\left\|\frac{\mathbf p(2.0)-\mathbf p(1.8)}{0.2}\right\|_2,\\
r^{\mathrm{long}}&=v_2-v_0,\qquad
r^{\mathrm{lat}}=\rho(2.0)-\rho(0).
\end{aligned}
\end{equation}
Predictions and records use the same recorded $v_0$, route geometry, and extraction rules. Longitudinal responses have units of m/s, while lateral responses have units of m. Route-relative offsets distinguish lateral adjustment from forward motion along a curved route.

The continuous response vector is defined as
\begin{equation}
\mathbf r_i =
\left(
\Delta v_i,\,
\Delta l_i,\,
\Delta\psi_i
\right),
\end{equation}
where $\Delta v_i=v_{2,i}-v_{0,i}$,
$\Delta l_i=\rho_i(2.0)-\rho_i(0)$, and $\Delta\psi_i$ denotes the heading change.
These components are extracted by the same trajectory-based response function for both
predictions and reference trajectories.

For the negative controls, only the predicted response vector is modified. The anchor-specific initial speed $v_{0,i}$ and all reference responses remain unchanged. The terminal speed is reconstructed from the modified longitudinal response as $v_{2,i}=v_{0,i}+\Delta v_i$.

\subsection{Evaluation Classes and SAT Mapping}
\label{app:action_mapping}

The auxiliary function $f_{\mathrm{SAT}}$ retains semantic action classes, such as turning, lane changing, and local avoidance. The evaluation function $f_{\mathrm{eval}}$ instead classifies the extracted trajectory response using fixed thresholds. SAT tokens are therefore not used directly as CECA reference labels. Table~\ref{tab:action_mapping} specifies the evaluation classes. Let $\Delta v=v_2-v_0$, $\Delta o=\rho(2.0)-\rho(0)$, and let $\Delta\psi$ be the change in route tangent direction between the projected anchor and terminal positions. Speed quantities are in m/s, offsets in meters, and angles in radians.

All predicted action classes are recomputed from the modified continuous responses using the same fixed thresholds. These include a stop-speed threshold of 0.20~m/s, gentle and strong
speed-change thresholds of 0.50 and 1.50~m/s, lateral thresholds of 0.35 and 0.75~m, and a
heading-change threshold of 0.20~rad. No action labels are directly permuted or manually
overridden in either negative control.

\begin{table}[!htbp]
\centering\small
\setlength{\tabcolsep}{4pt}
\caption{\textbf{Evaluation action classes and response thresholds.} Longitudinal stop/hold rules are applied before speed-change classes. SAT accelerate/decelerate families map to their gentle or strong evaluation classes. Semantic lateral tokens have no fixed one-to-one mapping to response classes.}
\label{tab:action_mapping}
\begin{tabularx}{\textwidth}{@{}l>{\raggedright\arraybackslash}p{0.23\textwidth}>{\raggedright\arraybackslash}X@{}}
\toprule
Direction & Evaluation class & Rule\\
\midrule
Longitudinal & Hold & $v_0\leq0.2$ and $v_2\leq0.2$\\
 & Stop & $v_0>0.2$ and $v_2\leq0.2$\\
 & Maintain speed & $-0.5<\Delta v<0.5$, after stop/hold cases\\
 & Gentle / strong acceleration & $0.5\leq\Delta v<1.5$ / $\Delta v\geq1.5$\\
 & Gentle / strong deceleration & $-1.5<\Delta v\leq-0.5$ / $\Delta v\leq-1.5$\\
\midrule
Lateral & Lane keeping & $|\Delta o|<0.35$ and $|\Delta\psi|<0.20$\\
 & Left / right steering & $0.35\leq|\Delta o|<0.75$ and $|\Delta\psi|<0.20$; direction follows the signed offset change\\
 & Sharp left / right steering & Remaining signed responses outside the lane-keeping and steering intervals\\
\bottomrule
\end{tabularx}
\end{table}

Route projection uses the nearest valid segment and requires at least two finite route points. A missing or degenerate reference route yields an unknown lateral class, which is excluded from the corresponding CECA target set. An invalid prediction on an eligible endpoint remains a failure. Each valid CECA endpoint is compared with its own recomputed reference class, independently of the other endpoint's class.

\section{Additional Diagnostic Results}
\label{app:extended_results}

Table~\ref{tab:response_diagnostics} supplements the main results with 2-second trajectory errors, absolute endpoint response errors, and CECA pair accuracy. Trajectory errors use the full valid test-anchor cohort. Conditional diagnostics use respective frozen paired subsets.

\begin{table}[!htbp]
\centering\small
\setlength{\tabcolsep}{4pt}
\renewcommand{\arraystretch}{1.05}
\caption{\textbf{Additional diagnostics for the completed runs.} Two-second ADE/FDE are in meters. Absolute response errors are in m/s longitudinally and m laterally. CECA$_{\mathrm{pair}}$ is the percentage of pairs with both endpoints correct under the stated aggregation. Check marks indicate training-only SAT.}
\label{tab:response_diagnostics}
\begin{fittable}
\begin{tabular*}{\textwidth}{@{\extracolsep{\fill}}lc*{6}{r}@{}}
\toprule
Method & SAT & \multicolumn{2}{c}{2-second error$\downarrow$} & \multicolumn{2}{c}{Absolute response error$\downarrow$} & \multicolumn{2}{c}{CECA$_{\mathrm{pair}}\uparrow$}\\
\cmidrule(lr){3-4}\cmidrule(lr){5-6}\cmidrule(lr){7-8}
 & & ADE & FDE & Long. & Lat. & Long. & Lat.\\
\midrule
Drive-JEPA & \xmark & 0.367 & 0.727 & 0.6403 & 0.1143 & 31.04 & 71.77 \\
Drive-JEPA & \cmark & 0.276 & 0.597 & 0.6276 & 0.0989 & 34.08 & 73.29 \\
\midrule
DriveLaW & \xmark & 0.224 & 0.432 & 0.5086 & 0.1727 & 17.60 & 70.10 \\
DriveLaW & \cmark & 0.216 & 0.429 & 0.4726 & 0.1787 & 22.97 & 73.07 \\
\midrule
WoTE & \xmark & 0.346 & 0.738 & 0.5549 & 0.2166 & 44.09 & 51.42 \\
WoTE & \cmark & 0.347 & 0.726 & 0.5567 & 0.2119 & 44.87 & 56.06 \\
\bottomrule
\end{tabular*}
\end{fittable}
\end{table}

\subsection{Trajectory Accuracy and Conditional Responses}

DriveLaW achieves the lowest 2-second ADE/FDE, at 0.224/0.432~m, but has lower longitudinal CECA than WoTE, at 44.38\% versus 57.90\%. Drive-JEPA reports lower lateral CRDE than DriveLaW, at 0.1586 versus 0.2680~m, despite higher 2-second trajectory errors. The reported rankings therefore differ even when trajectory evaluation uses the same response horizon. Position accuracy, response-difference agreement, and action-class correctness describe different aspects of planning. 

SAT improves Drive-JEPA at both prediction horizons. It reduces 2-second ADE/FDE by 24.82\%/17.82\% and 6-second ADE/FDE by 15.73\%/18.04\%. Longitudinal and lateral CRDE and CECA also improve. These results show consistent gains across trajectory and conditional evaluation for this backbone, with larger relative trajectory improvements over the initial response window.

\subsection{Absolute Response Accuracy and Pair Success}

DriveLaW with SAT reduces lateral CRDE from 0.2680 to 0.2619~m, while its absolute lateral response error increases from 0.1727 to 0.1787~m. Better agreement between response differences therefore does not imply more accurate individual responses. A common response bias at both endpoints cancels in CRDE. In comparison, Drive-JEPA reduces both lateral CRDE and absolute lateral response error, from 0.1586 to 0.1484~m and from 0.1143 to 0.0989~m, respectively. The absolute companion distinguishes improved relative responses from improved endpoint accuracy.

CECA pair accuracy reveals differences in two-endpoint success. With SAT, Drive-JEPA's longitudinal endpoint accuracy increases from 54.48\% to 54.58\%, while pair accuracy increases from 31.04\% to 34.08\%. For WoTE, longitudinal endpoint accuracy decreases by 0.44 percentage points, but pair accuracy increases by 0.78 percentage points. Endpoint accuracy and pair success can therefore change differently. Their outer weights also differ, since they average environments and unordered environment pairs, respectively. These observations support reporting both scores when examining action correctness across environments. The resulting diagnostics describe conditional performance on the evaluated subsets and do not establish causal dependence on a key factor.